\documentclass[pdflatex,sn-aps]{sn-jnl}

\usepackage[version=4]{mhchem}
\usepackage{subfig}
\usepackage{graphicx}%
\usepackage{multirow}%
\usepackage{amsmath,amssymb,amsfonts}%
\usepackage{amsthm}%
\usepackage{mathrsfs}%
\usepackage[title]{appendix}%
\usepackage{xcolor}%
\usepackage{textcomp}%
\usepackage{manyfoot}%
\usepackage{booktabs}%
\usepackage{algorithm}%
\usepackage{algorithmicx}%
\usepackage{algpseudocode}%
\usepackage{listings}%
\usepackage{tikz}%

\theoremstyle{thmstyleone}%
\theoremstyle{thmstyletwo}%

\theoremstyle{thmstylethree}%

\begin{document}

\title[Article Title]{AI-Guided Molecular Simulations in
VR: Exploring Strategies for Imitation Learning in
Hyperdimensional Molecular Systems}


\author*[1]{\fnm{Mohamed} \sur{Dhouioui}}\email{mohamed.dhouioui@usc.es}

\author[1]{\fnm{Jonathan} \sur{Barnoud}}

\author[1]{\fnm{Rhoslyn} \sur{Roebuck Williams}}

\author[1]{\fnm{Harry} \sur{J. Stroud}}

\author[2]{\fnm{Phil} \sur{Bates}}

\author[1]{\fnm{David} \sur{R. Glowacki}}

\affil*[1]{\orgdiv{IRL}, \orgname{CiTIUS Centro Singular de Investigación en Tecnoloxías Intelixentes}, \orgaddress{\city{Santiago de Compostela}, \country{Spain}}}

\affil[2]{\orgname{University of Bristol}, \orgaddress{\city{Bristol},\country{United Kingdom}}}


\abstract{Molecular dynamics (MD) simulations are a crucial computational tool for researchers to understand and engineer molecular structure and function in areas such as drug discovery, protein engineering, and material design. Despite their utility, MD simulations are expensive, owing to the high dimensionality of molecular systems. Interactive molecular dynamics in virtual reality (iMD-VR) has recently emerged as a `human-in-the-loop' strategy for efficiently navigating hyper-dimensional molecular systems. By providing an immersive 3D environment that enables visualization and manipulation of real-time molecular simulations running on high-performance computing architectures, iMD-VR enables researchers to reach out and guide molecular conformational dynamics, in order to efficiently explore complex, high-dimensional molecular systems. Moreover, iMD-VR simulations generate rich datasets that capture human experts’ spatial insight regarding molecular structure and function. This paper explores the use of researcher-generated iMD-VR datasets to train AI agents via imitation learning (IL). IL enables agents to mimic complex behaviours from expert demonstrations, circumventing the need for explicit programming or intricate reward design. In this article, we review IL across robotics and Multi-agents systems domains which are comparable to iMD-VR, and discuss how iMD-VR recordings could be used to train IL models to interact with MD simulations. We then illustrate the applications of these ideas through a proof-of-principle study where iMD-VR data was used to train a CNN network on a simple molecular manipulation task – namely, threading a small molecule through a nanotube pore. Finally, we outline future research directions and potential challenges of using AI agents to augment human expertise in navigating vast molecular conformational spaces.}

\keywords{molecular dynamics, virtual reality, imitation learning,}



\maketitle

\section{Introduction}\label{sec1}

Molecular dynamics (MD) simulations are a powerful tool for studying the structure, dynamics, and interactions of molecular systems. However, generating conformational ensembles and sampling rare events, (e.g., protein-ligand binding or protein conformational changes), remains challenging due to high computational costs and the complexity of the associated energy landscapes \cite{Saunders2017A}.
Interactive molecular dynamics in virtual reality (iMD-VR) has recently emerged as a promising approach to address these challenges by leveraging human intuition during real-time MD simulations within an immersive 3D environment \cite{Walters2022}.
In iMD-VR, users can directly manipulate and steer molecular systems using natural hand motions, applying forces to drive physically-relevant rare events such as conformational changes and ligand binding/unbinding \cite{oconnor_interactive_2019}.
This human-in-the-loop approach leverages the human's innate ability for 3D spatial reasoning and manipulation, enabling a user to intuitively explore complex molecular landscapes.
Recent studies have demonstrated the efficacy of iMD-VR in recreating crystallographic binding poses for protein-ligand systems \cite{Deeks2020, Deeks2020Interactive, Deeks2023}  and generating important reactive pathways \cite{Shannon2021}.
These interactive simulations capture valuable conformational data that can be challenging to obtain through conventional MD alone, thus offering new opportunities for applications such as training machine learning and investigating reaction mechanisms.

Imitation learning (IL) is a machine learning paradigm where algorithms extract knowledge from expert demonstrations by either directly imitating the desired behaviour or learning the expert's underlying reward function \cite{Zheng}. This approach allows machines to learn and replicate human actions, making it particularly valuable when manually designing reward functions is challenging, such as in noisy environments where it is easier to demonstrate desired behaviour than to engineer it. Learning from demonstration is often used interchangeably with IL in the literature \cite{Hussein, Zheng}, though some work distinguishes between offline approaches using pre-collected demonstrations versus interactive IL, where learners can actively query experts for demonstrations during training. The literature systematizes IL approaches primarily based on the type of information available from expert demonstrations\cite{gavenski2024surveyimitationlearningmethods}. Learning from Demonstration (LfD) represents the traditional approach that involves both action and state supervisions, providing complete expert guidance. In contrast, Learning from Observations (LfO) is a more practical variant that only has access to state-only demonstrations, making it more challenging due to incomplete expert guidance but enabling the use of previously inapplicable resources such as videos.
This approach has been particularly influential in the field of robotics, where it has been used to teach robots complex tasks without the need for explicit programming. By observing human demonstrations, robots can learn to perform a variety of actions, ranging from simple manipulations to complex, multi-step procedures \cite{Schaal1999,Chella2018, Hua2021, Zare2023ASO}. Learning from observation differs from other types of machine learning such as reinforcement learning, where an explicit reward function needs to be defined in advance or fine-tuned during training. IL offers the ability to learn a mapping of observations to actions performed by an expert in demonstrations, making it particularly well-suited to domains for which specifying a reward function is challenging or where human expertise can be leveraged.
The versatility of IL has also sparked interest in its application beyond robotics \cite{Leinen2020}, such as in MD  \cite{MT-MOL-GPT}, where it could potentially streamline the process of simulating and understanding complex molecular interactions.

A policy is a mapping from observations or states of the environment to actions that the agent should take. IL often requires a large number of demonstrations to effectively learn it, especially for complex tasks. Collecting a sufficiently large and diverse dataset of human demonstrations can be challenging for various reasons \cite{Zare2023ASO}. One challenge is that human behaviour is often multi-modal---there are many valid ways to perform a task. Standard IL approaches may average out these modes and learn a sub-optimal policy. Capturing and replicating diverse human behaviours is an open challenge \cite{jia2024towards}, mainly because robustly capturing all relevant aspects of human demonstrations can be difficult due to sensor limitations, occlusions, etc \cite{Maadi2021-dv}. This is especially true when collecting data outside of lab settings `in the wild'. Therefore there is a growing need for more large-scale, open datasets of human demonstrations on standardized tasks in order to facilitate reproducible research and benchmark IL algorithms \cite{jia2024towards, Webb2021}.

Virtual reality (VR) presents a novel and immersive platform for enhancing the capabilities of IL \cite{Zare2023ASO, Jung2021-ql}, where VR is combined with high-performance computing to provide an interactive environment in which researchers can manipulate molecular structures in real-time. The intuitive and engaging nature of VR could transform how scientists interact with molecular simulations, making it easier to collect data, hypothesize, and test the dynamics of molecular systems \cite{Deeks2020, Seritan2021}.

This paper aims to review IL's current applications in various domains, including key concepts like BC and GAIL. Particular focus is given to its potential in iMD-VR. We aim to provide a comprehensive review of the existing literature on IL, identify the benefits and challenges associated with its use, and propose innovative ways VR could serve as a data creation and collection platform for MD. By bridging the gap between IL and iMD-VR, we hope to open new avenues for research and application in the field of interactive MD, ultimately contributing to advancements in scientific understanding and technological development.

\section{Virtual reality for molecular simulations}\label{sec2}
\subsection{Molecular vizualisation}\label{subsec2}

Virtual reality (VR) provides researchers with natural, intuitive 3D interfaces to view and interact with complex molecular structures in a way that is not facilitated with traditional 2D interfaces. This can enhance the researcher's understanding of complex 3D molecular arrangements and interactions, which is essential for insight. Furthermore, VR can enhance scientific collaboration by providing shared virtual environments, which are accessible over the internet, thus enabling collaboration across physical distances. 

There exist several programs for the visualization of molecular simulations in VR, e.g., UnityMol\cite{UnityMol}, and the commercial software Nanome\cite{NaNome2023}. These programs provides a collaborative virtual environment in which users can visualize and manipulate molecular structures in stereoscopic 3D. Researchers can analyse the spatial arrangement of molecules, measure distances between atoms, and dock ligands into protein binding pockets using natural hand gestures \cite{Kneller2021}. Other examples of software include ProteinVR \cite{ProteinVR2020}, a web-based application that works across desktop, mobile, and VR platforms to visualize structural biology in 3D, and Molecular Rift \cite{Norrby2015-wz}, which provides controller-free manipulation of molecules using intuitive hand gestures.

By immersing users in 3D virtual environments, VR enables intuitive exploration of complex biomolecular systems that traditional 2D screens can not facilitate.
The application of VR to molecular visualisation unlocks several key benefits. First, it provides researchers with natural, intuitive 3D interfaces to view and interact with complex molecular structures. Second, VR enables collaborative molecular modelling in shared virtual environments, enhancing scientific teamwork. Finally, the stereoscopic depth perception and wide field of view in VR leads to an enhanced spatial understanding of 3D molecular arrangements \cite{Crossley-Lewis2023-jk}.

\subsection{Interactive molecular simulations}\label{subsec3}

Recent advancement in computational power and improving performance of graphical processing units has facilitated not only the visualization of molecules in VR, but also real-time interactivity. One example is NanoVer \cite{stroud_nanover_2025,oconnor_interactive_2019,jamieson-binnie_narupa_2020}, an open-source program developed by Glowacki et al. for performing iMD-VR. Using VR controllers, users can apply forces directly to MD simulations in real-time to drive important chemical events such as ligand binding and conformational changes.

Several aspects of NanoVer make it a suitable software for use in an iMD-VR+IL workflow. Firstly, NanoVer delivers quantitative chemical information on-the-fly such as potential energy, kinetic energy, and work done on the system during user interactions. These quantities provide physically motivated metrics with which to train IL models. Narupa \cite{oconnor_interactive_2019,jamieson-binnie_narupa_2020} (the precursor to NanoVer) has already been used to investigate reaction pathways \cite{Shannon2021}, ligand binding poses \cite{Deeks2020,Deeks2020Interactive}, and thermodynamic properties such as binding free energies \cite{Deeks2023}, which validates the use of this iMD-VR framework as a research-grade tool. To our best knowledge, this is the only open-source iMD-VR software that has proven utility in a chemistry research context. Furthermore, the program offers comprehensive documentation, active ongoing development, and compatibility with the latest commercial VR headsets. 

\begin{figure}[H]
    \centering
    \includegraphics[width=0.7\textwidth]{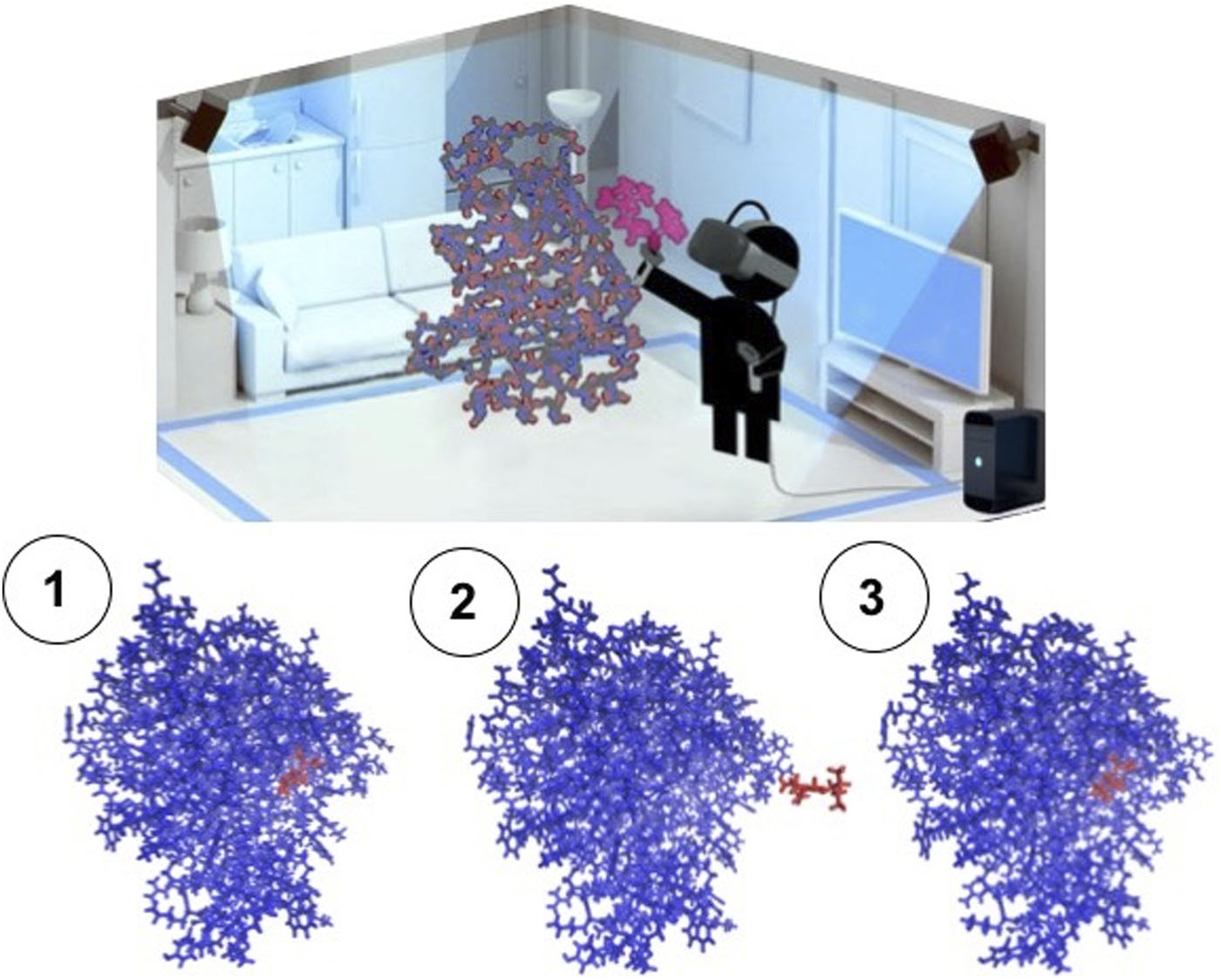}
    \caption{Showing an iMD-VR user docking and undocking the drug oseltamivir from the H7N9 neuraminidase protein (see text for further detail) \cite{oconnor_interactive_2019}}
    \label{fig:Figure1}
\end{figure}

An example of this is demonstrated in Figure \ref{fig:Figure1}. The top panel illustrates a researcher using iMD-VR to explore the binding and unbinding pathways of the drug oseltamivir, also known as tamiflu, (shown in magenta) with the H7N9 neuraminidase protein. The bottom panel shows snapshots of the molecular system at three time-points in the simulation: (1) oseltamivir bound to the active site of neuraminidase, (2) the molecular system after the researcher has undocked oseltamivir from the binding pocket of neuraminidase, and (3) the protein-ligand complex, where oseltamivir has been re-docked by the researcher after interactively exploring potential binding modes. 

\subsection{Molecular data structures in iMD-VR }\label{subsec4}

In essence, an MD simulation is a time series consisting of a set of frames of the atomic positions of a molecular system, which can be viewed frame-by-frame as a `trajectory' (an example is shown in Table \ref{tab:table1}). In iMD-VR, these trajectories are generated and visualized on the fly. Each frame contains information about the system (e.g. the temperature and energy), and the atoms contained within it (e.g. their position, velocities and element type). The iMD-VR program NanoVer uses a key-value system to store and communicate the data associated with these frames.
\begin{table}[h]
\caption{Dataframe of the first frame from a recording for the nanotube task}
    \label{tab:table1}
    \begin{tabular}{cccc}
    \toprule
       atom name & frame index & coordinates & user forces\\
    \midrule
        C1 & 0 & [9.725553, 14.941643, 14.158468] & [0.0, 0.0, 0.0]\\
        C2 & 0 & [10.063371, 15.170232, 12.954147] & [0.0, 0.0, 0.0]\\
        C3 & 0 & [11.367319, 15.154369, 12.419062] & [0.0, 0.0, 0.0]\\
        C4 & 0 & [11.99453, 16.465868, 12.049124] & [0.0, 0.0, 0.0]\\
        ... & ... & ... & ...\\
        H4 & 0 & [7.0092716, 18.310032, 12.723206] & [0.0, 0.0, 0.0]\\
    \botrule
    \end{tabular}
\end{table}
NanoVer streams data in real-time in the form of two dictionaries: (a) the frames, containing information about the simulation, such as the atomic positions, velocities, energies, and forces; and (b) the `shared state', consisting of synchronized information about avatar positions and user interactions such as controller positions and button pushes. The NanoVer server uses these streams to manage and process how users interact with MD simulation in VR. Moreover, it can record these data, thus enabling post-hoc analysis and playback of sessions. These recordings can be loaded onto the server, which then sends the recorded streams (synchronized using timestamps) to the clients as if they were real-time simulation streams. In this case, NanoVer acts purely as a molecular visualizer, affording the user control over the position/rotation/scale of the simulation and providing typical playback features (play/pause/restart). During playback, the user cannot apply forces to the molecular system. NanoVer recordings can be imported and analysed using a Python script with the MDanalysis package\cite{MDanalysis1, MDanalysis2}.

NanoVer offers developers and users the possibility of plugging in a selection of MD simulation software. One of the prototypical examples used for NanoVer demonstrations is the simulation of a methane molecule and a carbon nanotube, a molecular system that memics biomolecular channels that act as molecule-selective filters\cite{kalra_methane_2004}.

\begin{figure}[H]
    \centering
    \includegraphics[width=0.8\textwidth]{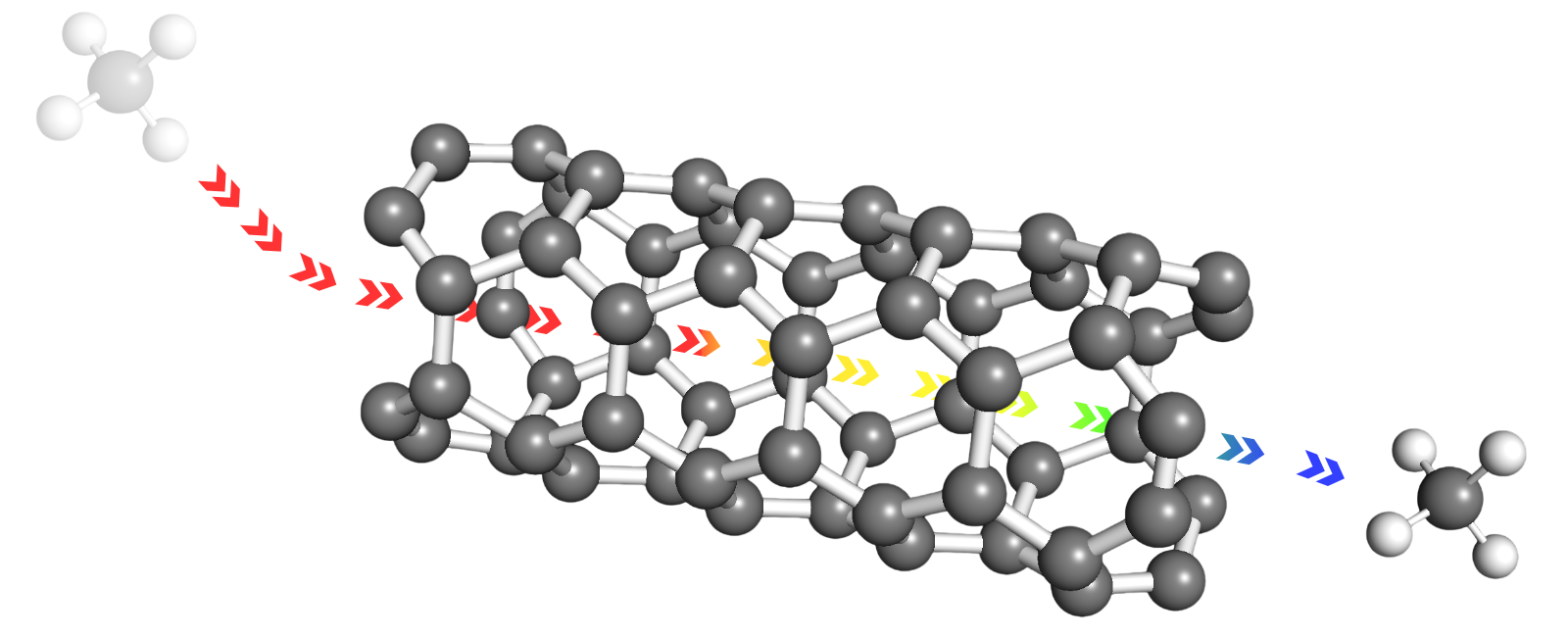}
    \caption{Trajectory depicting threading of a Methane molecule through a carbon Nanotube}
    \label{fig:Nanotube}
\end{figure}

In this simulation, users can thread a methane through a nanotube (Figure~\ref{fig:Nanotube}). 
The molecular system comprises 65 atoms: 60 carbons for the nanotube (labeled C1--C60), and 1 carbon and 4 hydrogens for methane (labeled C61 and H1--H4). Table~\ref{tab:table1} shows an example of the data collected for this system. Such systems, where users try to reach a specific goal (e.g. achieve a specific conformation, follow specific reaction coordinates, etc.), will be referred to as ``molecular tasks''.

 The data frame shown in Table \ref{tab:table1} has 4 columns: atom name, frame index, coordinates, and user forces. Here, \textit{atom name} is the atom's label, \textit{frame index} is the simulation frame index, \textit{coordinates} contain the (x,y,z) positions in nanometres, and \textit{forces} contain the (x,y,z) components of the forces in kilojoules per mole per nanometre (Fx, Fy, Fz) applied by the user on the specified atom.

\begin{figure} [H]
    \centering
    \includegraphics[width=0.7\textwidth]{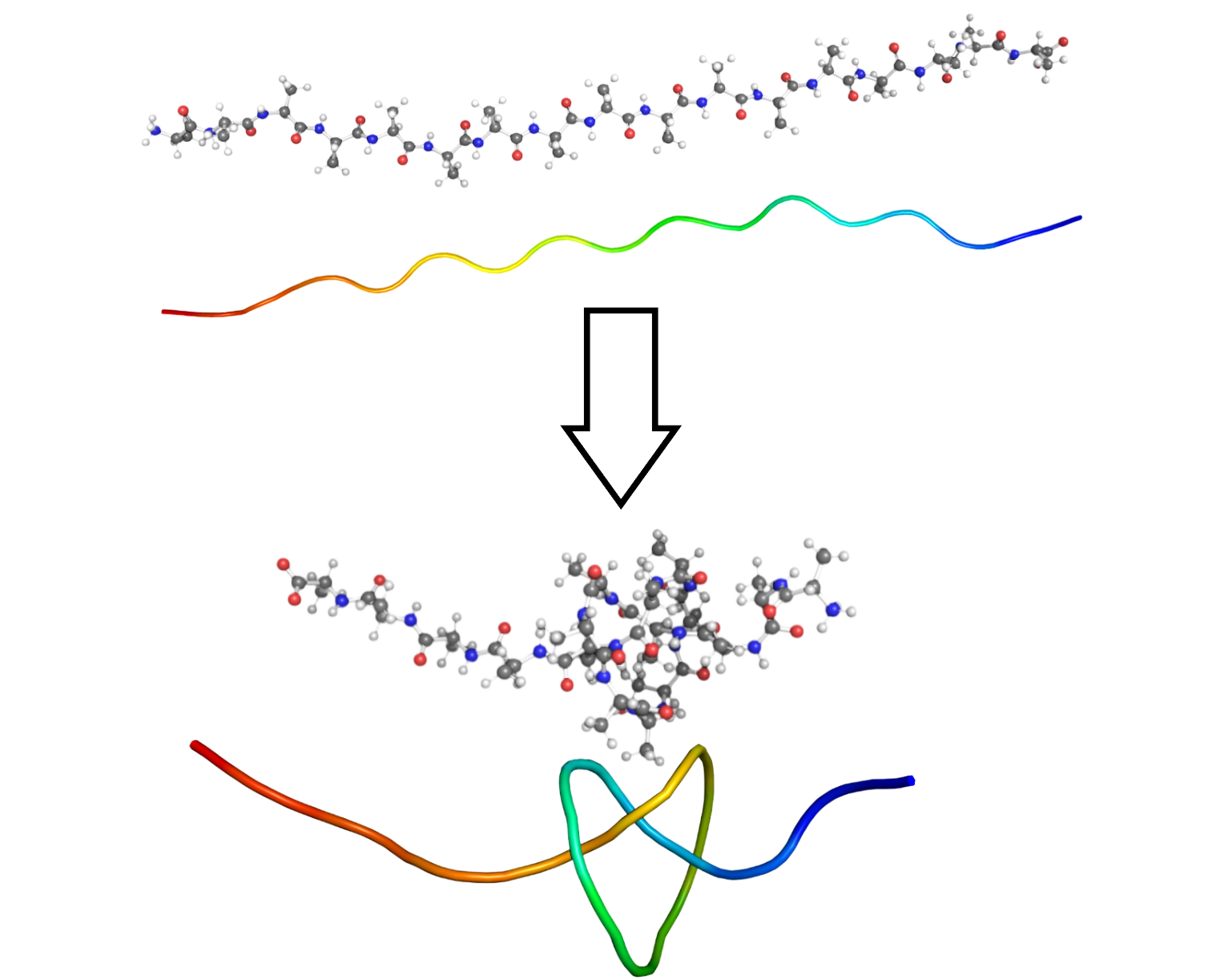}
    \caption{Knot tying task in 17 Alanine molecule}
    \label{fig:alanine}
\end{figure}

A more sophisticated molecular task is the one shown in Figure \ref{fig:alanine}. This task involves tying a knot in a 17-alanine molecule. The user is presented with the molecule in native linear form and asked to form a knot by manipulating both ends. The systems illustrated in Figs \ref{fig:Nanotube} and \ref{fig:alanine} are demonstration systems involving only small molecules. The software can be used on larger, more complex molecular systems such as protein systems.

\section{Imitation Learning in Agents and Multiagent systems}\label{sec3}

\subsection{Recent works in literature}\label{subsec6}
The recent surge of research on IL \cite{Zare2023ASO,correia2023survey,Hussein,gavenski2024surveyimitationlearningmethods} where policies are acquired by mimicking expert demonstrations instead of optimizing explicit reward definitions has generated a rich methodological landscape and a broad spectrum of applications. To orient this work, the literature review is structured into two subsections. The first contrasts the use of IL in single-agent and multi-agent systems, which studies on one side a solitary learner acting in an otherwise stationary environment, and on the other side multi-agent IL, where several learners must coordinate and reason over multiple interacting components; recent advances highlight how complex behaviours can emerge from simpler local interactions. This is particularly relevant for IL in iMD-VR to understanding how coordinated molecular manipulations can lead to desired conformational changes or structural transitions. The second subsections examine why IL has become a cornerstone of modern robotics: surveys and empirical studies show how it delivers sample-efficient, safety-aware skill acquisition for tasks ranging from autonomous driving to dexterous manipulation. The spatial reasoning capabilities developed in robotics are particularly relevant to molecular systems to developed imitation learning approaches for complex manipulation tasks that require understanding three-dimensional spatial molecular relationships. Together, these perspectives lay the conceptual and practical groundwork for leveraging insights from both algorithmic strands to inform the application of IL within iMD-VR. Both robotics and MAS address challenges that are central to iMD-VR applications: handling high-dimensional state spaces, learning from expert demonstrations, coordinating multiple interacting components, and achieving robust performance under uncertainty. The high-dimensional sampling problem that makes molecular dynamics computationally expensive parallels the high-dimensional control challenges in both robotics manipulation and multi-agent coordination. The human-in-the-loop aspect of iMD-VR also benefits from both domains - robotics provides frameworks for learning from human demonstrations and teleoperation, while MAS offers insights into human-agent interaction and coordination strategies.

By reviewing both robotics and MAS applications, we can leverage the complementary strengths of both fields: robotics' expertise in manipulation and spatial reasoning, and MAS's expertise in coordination and distributed control, creating a more comprehensive foundation for advancing molecular manipulation capabilities in virtual reality environments.

Whereas in single-agent systems, IL algorithms operate by training an individual agent to mimic expert demonstrations, focusing on state-action mapping without considering interactions with other learning entities, in multi-agent systems, imitation learning becomes significantly more complex as algorithms must simultaneously learn individual agent policies and coordination mechanisms from demonstrations of multiple interacting experts. Seeing that NanoVer\cite{stroud_nanover_2025} supports multi-user interactions and enables multiple persons to be in the same VR molecular environment, this multi-user capability directly aligns with multi-agent IL principles, as users can:
\begin{itemize}
    \item \textbf{Generate Coordinated Demonstrations:} Expert molecular manipulation teams can provide demonstrations of collaborative molecular dynamics tasks, showing how different users coordinate to achieve complex molecular manipulations.
    \item \textbf{Learn Coordinated Policies:} Multi-agent IL algorithms can learn from these multi-user demonstrations to understand both individual manipulation strategies and the coordination patterns that emerge during collaborative molecular exploration.
\end{itemize}
In recent years, IL in agents and multi-agent systems has seen significant advancements. One notable contribution is the introduction of Multi-agent Inverse Factorized Q-learning (MIFQ) \cite{bui2024inverse}, a novel algorithm that employs mixing networks to aggregate decentralized Q functions for centralized learning and uses hyper-networks to generate weights for mixing networks \cite{zhang2024inverse}. This approach has demonstrated superior performance compared to baseline algorithms in various multi-agent environments, including SMACv2 \cite{SMACv2}, Gold Miner \cite{Miner}, and Multi Particle Environments. MIFQ enables efficient and stable learning in cooperative multi-agent settings.

Another significant development in the field focuses on scaling laws for IL in single-agent games. Paine et al. \cite{paine2023scaling} investigate the impact of scaling up model and data size on IL performance, particularly in Atari games and NetHack \cite{NetHack}. By using Behavioural Cloning (BC) to imitate expert policies, they reveal that IL loss and mean return follow clear power law trends with respect to FLOPs. Importantly, loss and mean return are highly correlated, indicating that improvements in loss predictably translate to improved performance. Their research demonstrates that scaling up model and data size can provide significant improvements in agent performance, with the scaled-up approach surpassing prior state-of-the-art by 1.5x for NetHack.

In the realm of multi-agent systems, the Multi-Agent Adversarial Interaction Priors (MAAIP) \cite{MAAIP} approach adapts Multi-Agent Generative Adversarial IL (MAGAIL) for modelling interactions between agents. Younes et al. \cite{MAAIP} introduce new objectives for training the system and modelling self and opponent observations separately. MAAIP has proven effective for learning interactive behaviours between multiple agents and can be applied to scenarios where agents need to adapt to each other's actions. This approach demonstrates potential for improving IL in competitive or cooperative multi-agent settings.

These recent advancements collectively highlight the importance of scaling, efficient centralized learning in decentralized execution settings, and modelling agent interactions for improved performance in complex environments. As the field of IL continues to evolve, these contributions pave the way for more sophisticated and effective agent behaviours in both single-agent and multi-agent systems.

\subsubsection{Manipulation tasks}\label{subsec7}

IL has found a wide range of applications in robotics, demonstrating its versatility and effectiveness in enabling robots to perform complex tasks. The following subsections delve into three primary areas where IL has been significantly applied: manipulation tasks, locomotion and navigation, and human-robot interaction.

Manipulation tasks involve robots handling, moving, or altering the state of objects in their environment. In recent years, IL has been instrumental in teaching robots to perform such tasks with precision and adaptability\cite{Ravichandar2020}.
For instance, VIOLA \cite{Zhu2022}, a novel IL approach that was implemented and deployed into a real-life robot, outperforms state-of-the-art methods by 45.8\% in success rate. This is achieved through the use of a pre-trained vision model which is put into a transformer-like architecture. The authors created a policy to detect task-driven relevant regions for action mapping.
Another novel hybrid imitation learning (HIL) framework combines behaviour cloning (BC) and state cloning (SC) methods to efficiently learn manipulation tasks like pick-and-place and stacking\cite{Jung2021-ql}. This approach has been shown to significantly improve training efficiency and policy flexibility, demonstrating a performance improvement and faster training time compared to pure BC methods. Hua et al. \cite{Hua2021} emphasize the efficiency of learning from good samples and the potential for combining reinforcement learning mechanisms to improve the speed and accuracy of IL. They specifically addressed the application of IL in robot manipulation by observing expert demonstrations, which can be generalized to other unseen scenarios.

\subsubsection{Locomotion and navigation}\label{subsubsec1}

In robotics, locomotion and navigation are two fundamental aspects that enable robots to move and operate within their environments effectively. These concepts are crucial for the development of autonomous systems that can perform a wide range of tasks, from simple delivery services to complex exploration missions. Locomotion refers to the various methods that robots use to move from one place to another. This movement can be achieved through different mechanisms, depending on the robot's design and the environment it is intended to operate in. Navigation involves the process by which a robot determines its position in the environment and plans a path to reach a specific destination.

Seo et al. \cite{Seo2023} introduce PRELUDE, a hierarchical learning framework designed to enhance the navigation and locomotion capabilities of quadrupedal robots in dynamic and cluttered environments.
The framework divides the problem into two levels: high-level navigation decision-making and low-level gait generation. The high-level controller is trained using IL from human demonstrations collected with a steerable cart, enabling the robot to acquire complex navigation behaviours. The low-level gait controller is trained through reinforcement learning, allowing the discovery of versatile gait patterns through trial and error. The effectiveness of PRELUDE is demonstrated through simulations and hardware experiments, showing significant improvements over state-of-the-art reinforcement learning methods in terms of success rate and travel distance in various environmental conditions\cite{Seo2023}.
This work exemplifies the application of IL in robotics, particularly in the development of autonomous systems capable of agile and adaptive movement in real-world scenarios.

\subsubsection{Human robot interaction}\label{subsubsec2}
Human interaction tasks in robotics involve robots engaging in various forms of social interaction and cooperation with humans to achieve shared goals. These tasks encompass direct physical interaction, such as assisting with lifting objects or providing physical therapy, as well as collaborative interaction, where robots and humans work together to complete tasks like assembling products on a manufacturing line. Remote interaction, where humans control or collaborate with robots from a distance, also falls under the umbrella of human-robot interaction tasks. The ultimate goal in human-robot interaction tasks is to achieve natural, efficient, and safe interactions as robots work with humans across various domains.
Mehta et al. \cite{Mehta2023} introduce a learning formalism that unifies approaches for physical human-robot interaction by incorporating demonstrations, corrections, and preferences. It represents a comprehensive approach to learning from human interactions, aiming to improve robot adaptability and performance in collaborative tasks. This framework is designed to learn without making assumptions about the tasks the human wants to teach the robot. The key insight to take away from this study by Mehta et al. is that physical human-robot interaction can be a rich source of information for teaching robots, and that by leveraging all available forms of interaction—kinesthetic guidance (demonstrations), adjustments to the robot's motion (corrections), and evaluative feedback (preferences)—a more robust and flexible learning system can be developed.
The authors propose a two-step algorithm that first learns a reward model from scratch by comparing the human's input to nearby alternatives and then applies constrained optimization to map the learned reward into a robot trajectory. This process is iterative and allows for real-time updates based on the human's feedback, which can be provided in any order and combination. The approach was validated through simulations and a user study, demonstrating that it can more accurately learn manipulation tasks from physical human interaction than existing baselines, especially when faced with new or unexpected objectives.\cite{Mehta2023}
They emphasize the importance of a unified learning approach that does not rely on predefined task features or reinforcement learning, thus enabling robots to learn new and unexpected tasks in real-time from physical interaction with humans. This has significant implications for the development of robots that can adapt to a wide range of tasks in shared human-robot environments, such as factories, homes, or healthcare settings, where safety and adaptability are paramount.

\subsection{Key concepts and techniques}\label{subsec8}
\subsubsection{Behavioural cloning}\label{subsubsec4}
Behavioural cloning (BC) is a straightforward approach that treats IL as a supervised learning problem\cite{Pomerleau1991}. Given a dataset of state-action pairs from expert demonstrations, BC directly learns a policy (mapping from states to actions) using regression or classification algorithms\cite{Sammut2011}. The policy is trained to minimize some loss function between the predicted and demonstrated actions on the training data. Based on the demonstration quality, BC is somewhat simple to implement since no extensive knowledge of the environmental dynamics is required. Being treated as supervised learning, a method that is very well studied, makes training BC algorithms computationally efficient.

Consider a dataset \(\mathcal{D} = \{(s_i, a_i)\}_{i=1}^M\) consisting of state-action pairs collected from an expert policy \(\pi^*\), where \(s_i\) represents the state and \(a_i\) the action taken by the expert in that state. The objective of BC is to learn a policy \(\hat{\pi}\) that approximates the expert policy \(\pi^*\) as closely as possible.

\begin{figure}[H]
    \centerline{\includegraphics[width=0.9\textwidth]{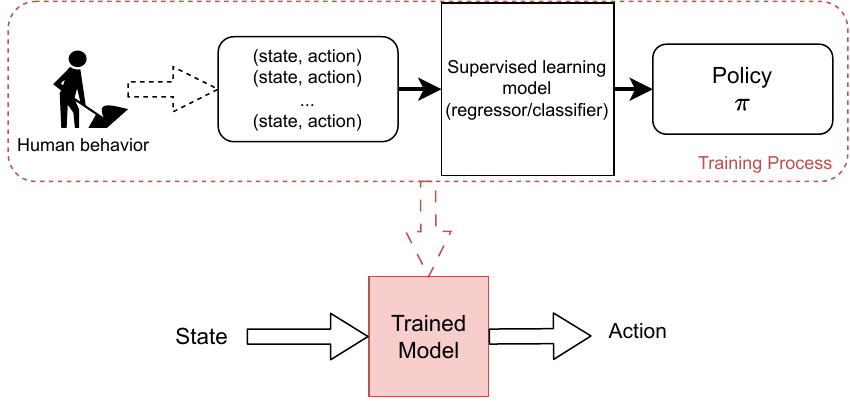}}
    \caption{Process of BC}
    \label{procstructfig2}
\end{figure}

Figure~\ref{procstructfig2} illustrates the process by which this is achieved, which involves:
\begin{enumerate}
\item{\textbf{Data Collection:}} Collect a dataset \(\mathcal{D}\) of state-action pairs \((s, a) \) by observing an expert performing the task.

\item{\textbf{Learning:}} Train a model \(\hat{\pi}\) on \(\mathcal{D}\) to learn the mapping from states to actions. This typically involves minimizing a loss function over the dataset. For discrete action spaces, a common choice is the negative log-likelihood (NLL) loss:
\begin{equation}
    \mathcal{L}(\pi, s, a^*) = -\ln \pi(a^*|s)
\end{equation}

   For continuous action spaces such as iMD-VR, the mean squared error (MSE) loss is often used:
   \begin{equation}
     \mathcal{L}(\pi, s, a^*) = \|\pi(s) - a^*\|^2 
   \end{equation}
\item{\textbf{Policy Output:}} After training, the learned policy \(\hat{\pi}\) can be used to perform the task, ideally replicating the expert's performance.
\end{enumerate}
\subsubsection{Inverse reinforcement learning}

Inverse Reinforcement Learning (IRL) is a method used to infer the reward function of an agent by observing its behaviour within an environment \cite{Russel1998, AndrewNg}. It assumes that we observe an agent following an unknown policy \(\pi^*\) and we want to infer the reward function \(R\) that this policy is optimizing. The problem is challenging because there are potentially many reward functions that could explain the observed behaviour, making IRL an ill-posed problem. IRL is often modelled as a Markov Decision Process (MDP) \cite{AndrewNg} where the goal is to determine what objectives or values the agent is optimizing for, given its observed actions.

To understand IRL, we first need to understand the framework in which it operates, which is the MDP. An MDP is defined by a tuple \((S, A, T, \gamma, R)\):

\begin{itemize}
    \item \(S\) is a set of states.
    \item \(A\) is a set of actions.
    \item \(T\) is the transition probability matrix, where \(T(s'|s, a)\) gives the probability of transitioning to state \(s'\) from state \(s\) after taking action \(a\).
    \item \(\gamma\) is the discount factor, which determines the present value of future rewards.
    \item \(R\) is the reward function, which assigns a scalar reward to each state (or state-action pair).
\end{itemize}

A policy \(\pi\) is a mapping from states to actions, and the goal in reinforcement learning is to find an optimal policy \(\pi^*\) that maximizes the expected sum of discounted rewards.

\begin{figure}
    \centerline{\includegraphics[width=0.9\textwidth]{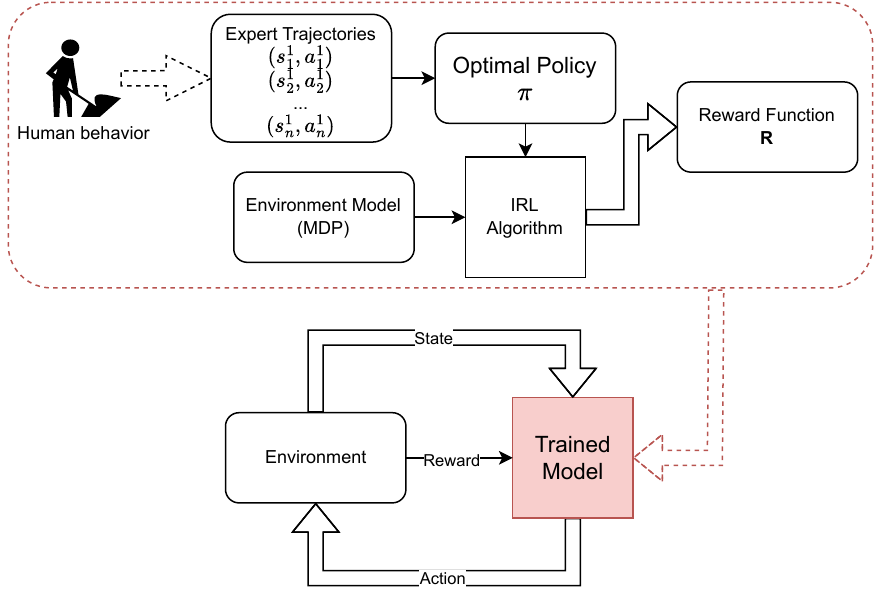}}
    \caption{Process of Inverse Reinforcement Learning (IRL)}
    \label{procstructfig3}
\end{figure}

Figure ~\ref{procstructfig3} demonstrates the general approach to IRL which involves the following steps:
\begin{enumerate}
    \item \textbf{Collecting Data:} Observe the behaviour of the expert agent and collect state-action trajectories.
    \item \textbf{Estimating the MDP:} Use the collected data to estimate the transition probabilities \(T\) and the initial state distribution.
    \item \textbf{Learning the Reward Function:} Infer a reward function \(R\) that would make the observed behaviour appear optimal.
\end{enumerate}

The mathematical formulation of IRL can be described as follows:

\begin{itemize}
    \item Given:
    \begin{itemize}
        \item A set of observed trajectories \(\tau = \{(s_1, a_1), (s_2, a_2), \ldots\}\) from an expert policy \(\pi^*\).
        \item An estimated MDP \((S, A, T, \gamma)\) without the reward function.
    \end{itemize}
    \item Find:
    \begin{itemize}
        \item A reward function \(R: S \times A \rightarrow \mathbb{R}\) such that the expert policy \(\pi^*\) is optimal for this reward function.
    \end{itemize}
\end{itemize}

One common approach to solving IRL is to use a linear approximation of the reward function, where \(R(s, a) = \theta^T \phi(s, a)\), and \(\phi(s, a)\) is a feature representation of the state-action pair. The parameters \(\theta\) are then learned by optimizing a likelihood function or by matching the feature expectations of the expert's policy.
Several algorithms have been proposed for IRL, including:
\begin{itemize}
    \item \textbf{Maximum Entropy IRL:} First introduced by Ziebart et al.\cite{Ziebart-2008} this method assumes that the expert behaves in a way that maximizes entropy, meaning that among all policies that could explain the expert's behaviour, the one that is least committed to unnecessary constraints is chosen.
    \item \textbf{Bayesian IRL:} This approach first introduced to IRL by Ramachandran et al.\cite{Ramchandran-2007} treats the reward function as a random variable and uses Bayesian methods to infer a posterior distribution over the reward function given the observed behaviour.
\end{itemize}

\subsubsection{Adversarial imitation learning}
IRL algorithms seen previously have a high computational complexity \cite{pmlr-v139-metelli21a}\cite{pmlr-v205-deka23a} since they require the execution of RL in inner loops \cite{Ho2016} (shown in Figure \ref{procstructfig3} as the IRL Algorithm box connection to Optimal Policy and meaning that for each candidate reward function, a complete RL problem to find the optimal policy is solved). Adversarial IL has been proposed as a solution to this computational challenge\cite{Ho2016}.
Generative Adversarial Imitation Learning (GAIL) is an adversarial IL algorithm that uses the framework of generative adversarial networks (GANs) to directly learn a policy from expert demonstrations, without needing to first learn a reward function as in IRL\cite{Ho2016}.

The key idea is to train a generator policy to produce trajectories that are indistinguishable from the expert trajectories, as judged by a discriminator network. This is formulated as a minimax game between the generator and discriminator\cite{Ho2016}:
\begin{equation}
    \min_\pi \max_D \mathbb{E}_{\pi}[\log D(s,a)] + \mathbb{E}_{\pi_E}[\log(1 - D(s,a))] - \lambda H(\pi)
\end{equation}

where:
\begin{itemize}
    \item \(\pi\) is the generator policy
    \item \(D\) is the discriminator
    \item \(\pi_E\) is the expert policy
    \item \(H(\pi)\) is an entropy regularization term
\end{itemize}

Let \(\rho_\pi(s,a)\) denote the occupancy measure, i.e. the distribution of states and actions encountered when navigating the environment with policy \(\pi\)\cite{Ho2016}.

GAIL seeks a policy whose occupancy measure matches the expert's:
\begin{equation}
    \rho_\pi(s,a) \approx \rho_{\pi_E}(s,a)
\end{equation}

It can be shown that finding a policy to minimize the Jensen-Shannon divergence between occupancy measures is equivalent to the following:
\begin{equation}
    \arg \min_\pi -H(\pi) + \psi_{GA}(\rho_\pi - \rho_{\pi_E})
\end{equation}

where \(\psi_{GA}\) is a convex regularizer with the form:
\begin{equation}
    \psi_{GA}(\rho_\pi) = \max_D \mathbb{E}_\pi[\log D(s,a)] + \mathbb{E}_{\pi_E}[\log(1 - D(s,a))]
\end{equation}

This leads to the GAIL objective in the first equation. The discriminator \(D\) is trained to distinguish expert vs policy state-action pairs, while the policy \(\pi\) is trained to maximize the discriminator confusion.

The GAIL algorithm alternates between training the discriminator and taking policy gradient steps:
\begin{enumerate}
    \item Sample trajectories \(\tau_i \sim \pi_{\theta_i}\) from current policy
    \item Update discriminator parameters to maximize:
    \(\mathbb{E}_{\tau_i}[\nabla_w \log D_w(s,a)] + \mathbb{E}_{\tau_E}[\nabla_w \log (1-D_w(s,a))]\)
    \item Take a policy gradient step using cost function \(\log D_w(s,a)\), e.g. using TRPO\cite{SchulmanLMJA15} (Trust Region Policy Optimization)
\end{enumerate}

The policy optimization uses a model-free RL algorithm like TRPO, using the discriminator output as the reward signal. Training continues until the policy performs well and the discriminator is unable to distinguish policy and expert state-action pairs.

GAIL leverages the expressive power of GANs to directly imitate expert demonstrations, without needing to recover a reward function explicitly as in inverse RL. The discriminator learns to distinguish expert data, providing a reward signal to optimize the policy to match the expert's occupancy measure. This enables the imitation of complex behaviours from a relatively small number of demonstrations.

\section{Strategies for applying IL to iMD-VR}

iMD-VR represents a paradigm shift in how researchers interact with and manipulate molecular systems, offering an intuitive and immersive approach to exploring conformational spaces. However, the manual nature of iMD-VR interactions, while valuable for gaining physical intuition about molecular systems, poses limitations in terms of scalability and reproducibility when generating large datasets of conformational pathways. The integration of IL with iMD-VR presents a promising solution to these challenges, as it enables the capture and reproduction of expert manipulation strategies in an automated fashion. By learning from demonstrations performed by experienced researchers in VR environments, IL algorithms can extract the underlying patterns and decision-making processes that guide successful conformational transitions, effectively creating a bridge between human intuition and computational efficiency. Furthermore, In iMD-VR, both training and inference can be conducted entirely within the virtual environment, eliminating the need to bridge the gap between simulation and physical reality, which is the case in robotics. This approach is more straightforward for several reasons.

The combination of IL and iMD-VR is particularly valuable for generating comprehensive datasets of conformational pathways, as it addresses several key limitations of traditional MD approaches. First, it allows for the systematic reproduction of rare events and complex transitions that might be difficult to sample using conventional simulation methods. Second, the learned models can generalize across similar molecular systems, potentially enabling the rapid exploration of conformational spaces for novel compounds based on previously learned manipulation strategies. Furthermore, the resulting datasets can serve as valuable training resources for developing more sophisticated machine learning models, creating a positive feedback loop that continuously improves our ability to predict and understand molecular behaviour. This synergistic approach not only accelerates the generation of conformational pathway datasets but also preserves the physical intuition and expertise embedded in human-guided explorations, making it an invaluable tool for advancing our understanding of molecular systems.

For instance, a pre-trained vision model in a transformer-like architecture, similar to VIOLA \cite{Zhu2022}, could be used to detect task-driven relevant regions of the nanotube and methane molecule for precise action mapping. A hybrid approach combining behaviour cloning and state cloning, inspired by HIL \cite{Jung2021-ql}, could efficiently learn the manipulation task of threading the molecule through the nanotube or learn to tie a knot by observing expert demonstrations in VR.
In similar implementations of IL, researchers have identified several significant challenges, including covariate shift \cite{Pomerleau1988}, causal misidentification \cite{deHaan2019}, and the copycat problem \cite{ChuanWen2020}. Each of these challenges undermines the effectiveness of IL algorithms, but the literature over the years has proposed innovative solutions to mitigate these issues, leading to more robust and generalizable models.

Covariate shift \cite{Pomerleau1988}, a prevalent issue where the training data distribution does not match the test data distribution, has been a focal point of concern. This mismatch leads to models that perform well on training data but fail to generalize to new, unseen environments. To combat this, IL techniques \cite{Jonathan2021} such as DAgger \cite{pmlr-v15-ross11a} (Dataset Aggregation) have been developed. These methods iteratively refine the training dataset by incorporating data collected under the policy currently being learned, thus aligning the training and test distributions more closely. Furthermore, IRL approaches that focus on learning the underlying reward function from expert demonstrations offer another avenue to address covariate shift. By concentrating on the reward structure rather than directly mimicking actions, these methods aim to achieve better generalization. Additionally, constrained IL \cite{chang2021mitigating} introduces constraints into the learning problem to prevent significant deviations from the expert policy, thereby reducing the impact of covariate shift.

Causal misidentification \cite{deHaan2019}, where models learn incorrect causations between observations and actions, poses another challenge. This issue can lead to models that make decisions based on spurious correlations, resulting in suboptimal or incorrect behaviour. Researchers have tackled this problem by applying causal inference techniques \cite{Gokul2022} to distinguish between causally relevant and irrelevant features. Moreover, structured IL methods \cite{ruan2023causal} that incorporate knowledge about the task or environment into the learning process help the model focus on the correct causal relationships, enhancing its decision-making capabilities.

The copycat problem \cite{ChuanWen2020}, characterized by models mimicking expert actions without understanding the underlying task structure, has also received attention. Solutions such as residual action prediction \cite{chuang2022resolving}, where the model predicts deviations from the expert's actions, encourage a deeper understanding of the task dynamics. Temporal regularization, which penalizes large changes in actions over time, further discourages models from blindly copying expert actions, promoting a more nuanced approach to learning from demonstrations.
\subsection{Proposed approach}
As an initial proof of concept we investigated the nanotube threading task as a BC challenge through a supervised learning framework, specifically employing a Convolutional Neural Network (CNN) architecture. CNNs have proven effective in dealing with spatial data where they show good performance in extracting hierarchical patterns. In our example, CNN will process spatial coordinates of all particles in the system as input, represented as 3D positional data for each simulation frame. The network's primary objective is to predict two critical parameters: the optimal interaction position vector and the corresponding force scale factor to be applied to the central carbon atom (C61) of the methane molecule. This formulation transforms the complex threading process into a time dependant regression problem, where the model learns the mapping between the instantaneous molecular configuration and the required force parameters that would facilitate efficient passage of methane through the carbon nanotube. By leveraging the CNN's ability to capture spatial hierarchies and local correlations in the input data, we aim to automate the threading process while maintaining physical consistency in the MD simulation. The predicted force parameters will be directly applied to guide the methane molecule's trajectory, potentially optimizing the threading efficiency compared to traditional methods.
\subsubsection{Network Design}
As an initial architecture we propose a model which employs a deep CNN structure comprising five convolutional layers with increasing feature dimensionality (32 to 512 channels). Each convolutional layer utilizes a kernel size of 3 with stride 1 and appropriate padding to maintain spatial dimensions. Max pooling operations with kernel size 2 and stride 2 are applied between convolutional layers to reduce spatial dimensionality while retaining salient features progressively. The choice of this architecture is set a an initial approach mirroring previous successful approaches. In future work, we intend to explore other approaches as highlighted in literature.
The architecture can be represented as:
\usetikzlibrary{arrows.meta,positioning,shapes.geometric}
\begin{figure}[H]
\centering
\begin{tikzpicture}[
    node distance=0.5cm,
    box/.style={rectangle,draw,minimum width=1.5cm,minimum height=0.5cm},
    arrow/.style={-{Stealth[length=3mm]},thick},
    every path/.style={rounded corners=5pt}
]

\node[box] (input) {Input(195, 1)};
\node[box] (conv1) [below=of input] {Conv1D(1→32)(195, 32)};
\node[box] (pool1) [below=of conv1] {MaxPool (97, 32)};
\node[box] (conv2) [below=of pool1] {Conv1D(32→64) (97, 64)};
\node[box] (pool2) [below=of conv2] {MaxPool (48, 64)};
\node[box] (conv3) [below=of pool2] {Conv1D(64→128) (48, 128)};

\node[box] (pool3) [right=3cm of input] {MaxPool (24, 128)};
\node[box] (conv4) [below=of pool3] {Conv1D(128→256) (24, 256)};
\node[box] (pool4) [below=of conv4] {MaxPool (12, 256)};
\node[box] (conv5) [below=of pool4] {Conv1D(256→512) (12, 512)};
\node[box] (pool5) [below=of conv5] {MaxPool (6, 512)};
\node[box] (fc1) [below=of pool5] {FC(3072→512) (512)};

\node[box] (fc2) [right=3cm of pool3] {FC(512→256) (256)};
\node[box] (fc3) [below=of fc2] {FC(256→128) (128)};
\node[box] (fc4) [below=of fc3] {FC(128→32) (32)};
\node[box] (fc5) [below=of fc4] {FC(32→4) (4)};
\node[box] (output) [below=of fc5] {Output (4)};

\draw[arrow] (input) -- (conv1);
\draw[arrow] (conv1) -- (pool1);
\draw[arrow] (pool1) -- (conv2);
\draw[arrow] (conv2) -- (pool2);
\draw[arrow] (pool2) -- (conv3);

\draw[arrow] (pool3) -- (conv4);
\draw[arrow] (conv4) -- (pool4);
\draw[arrow] (pool4) -- (conv5);
\draw[arrow] (conv5) -- (pool5);
\draw[arrow] (pool5) -- (fc1);

\draw[arrow] (fc2) -- (fc3);
\draw[arrow] (fc3) -- (fc4);
\draw[arrow] (fc4) -- (fc5);
\draw[arrow] (fc5) -- (output);

\draw[arrow] (conv3) -- ++(3cm,0) |- (pool3);
\draw[arrow] (fc1) -- ++(3cm,0) |- (fc2);

\end{tikzpicture}
\caption{CNN Architecture with Tensor Shapes}
\label{fig:cnn_architecture}
\end{figure}
The fully connected layers progressively reduce dimensionality from 3072 to the final output dimension of 4, representing the predicted interaction parameters. ReLU activation functions are employed throughout the network to introduce non-linearity.
\subsubsection{Training Methodology}
The training process utilized a comprehensive dataset derived from 20 expert-guided virtual reality recordings of successful methane-nanotube threading operations whith 80\% used for training and 20\% for validation. Each frame in the training set was processed independently, containing a total of 65 atomic positions: 60 positions from the carbon nanotube structure and 5 positions representing the methane molecule (one central carbon and four hydrogen atoms). The target variables for each frame consisted of a four-dimensional vector, comprising the three-dimensional interaction position $(x,y,z)$ and a scalar force scale parameter. The atomic positions were arranged as a one-dimensional sequence to facilitate the CNN's ability to extract spatial features and inter-atomic relationships. Training was conducted using the Adam optimization algorithm with an initial learning rate of 0.001, employing Mean Squared Error (MSE) as the loss function to minimize prediction errors. Model performance was evaluated using the R² score metric, which quantifies the proportion of variance in the target variables that is predictable from the atomic positions. This frame-by-frame approach ensures that the model learns the temporal independence of the threading process while capturing the essential spatial correlations between atomic positions.
\subsection{Initial Results}

\begin{figure}
    \centering
    \subfloat[CNN Training Loss]{\includegraphics[width=0.5\linewidth]{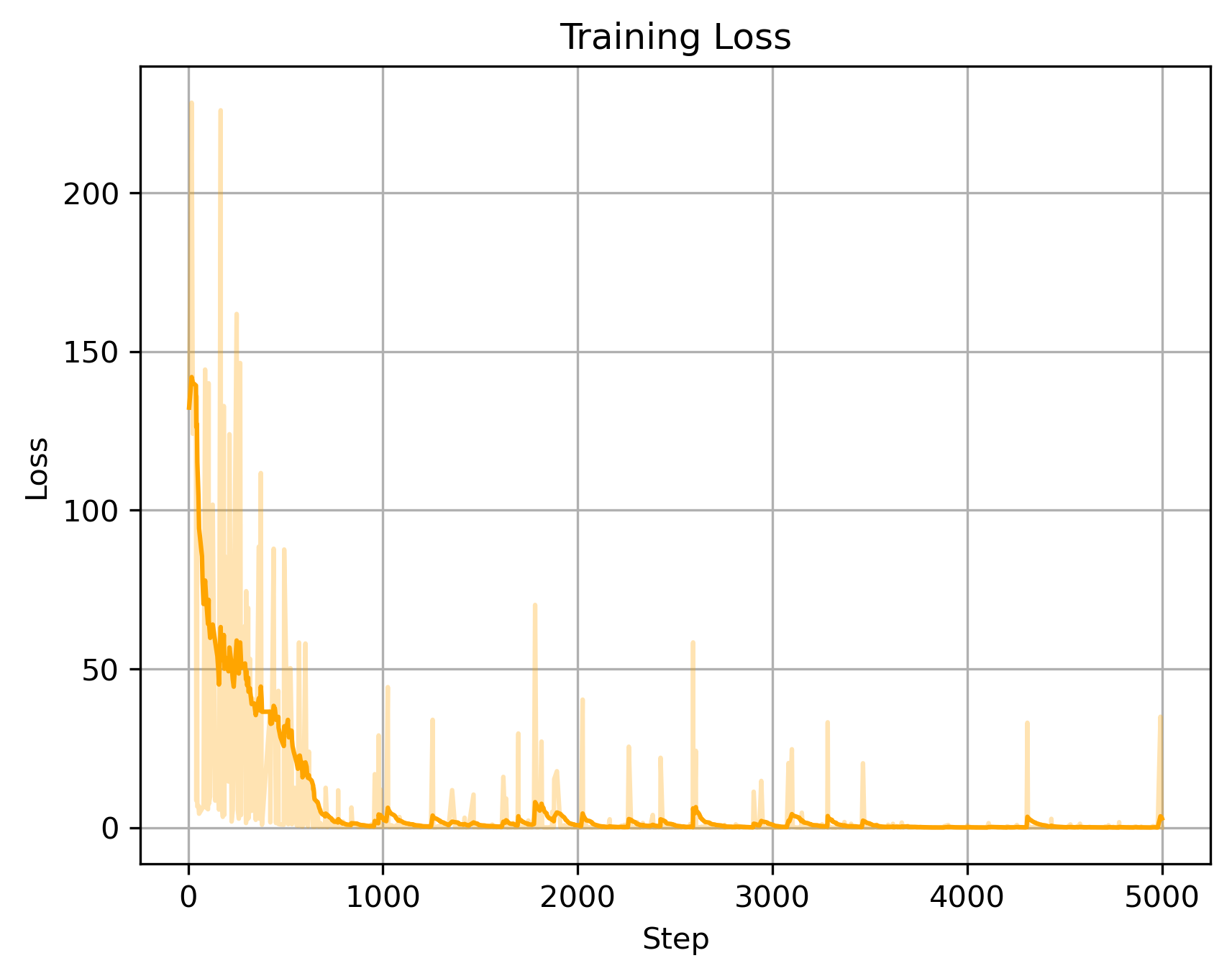}}
    \subfloat[CNN Validation Loss]{\includegraphics[width=0.5\linewidth]{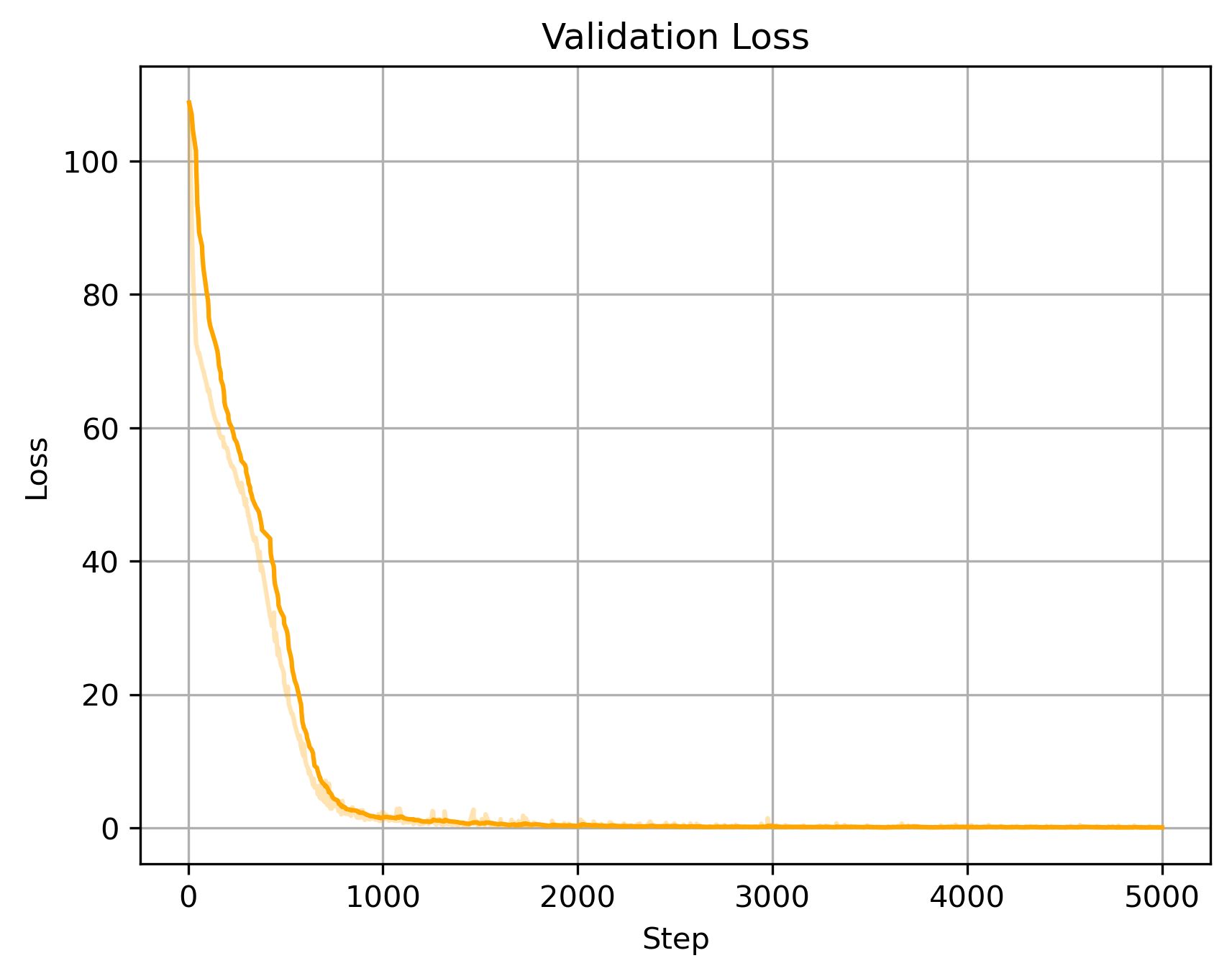}}
    \caption{CNN loss evolution plots}
    \label{loss_plots}
\end{figure}

\begin{figure}
    \centering
    \includegraphics[width=0.5\linewidth]{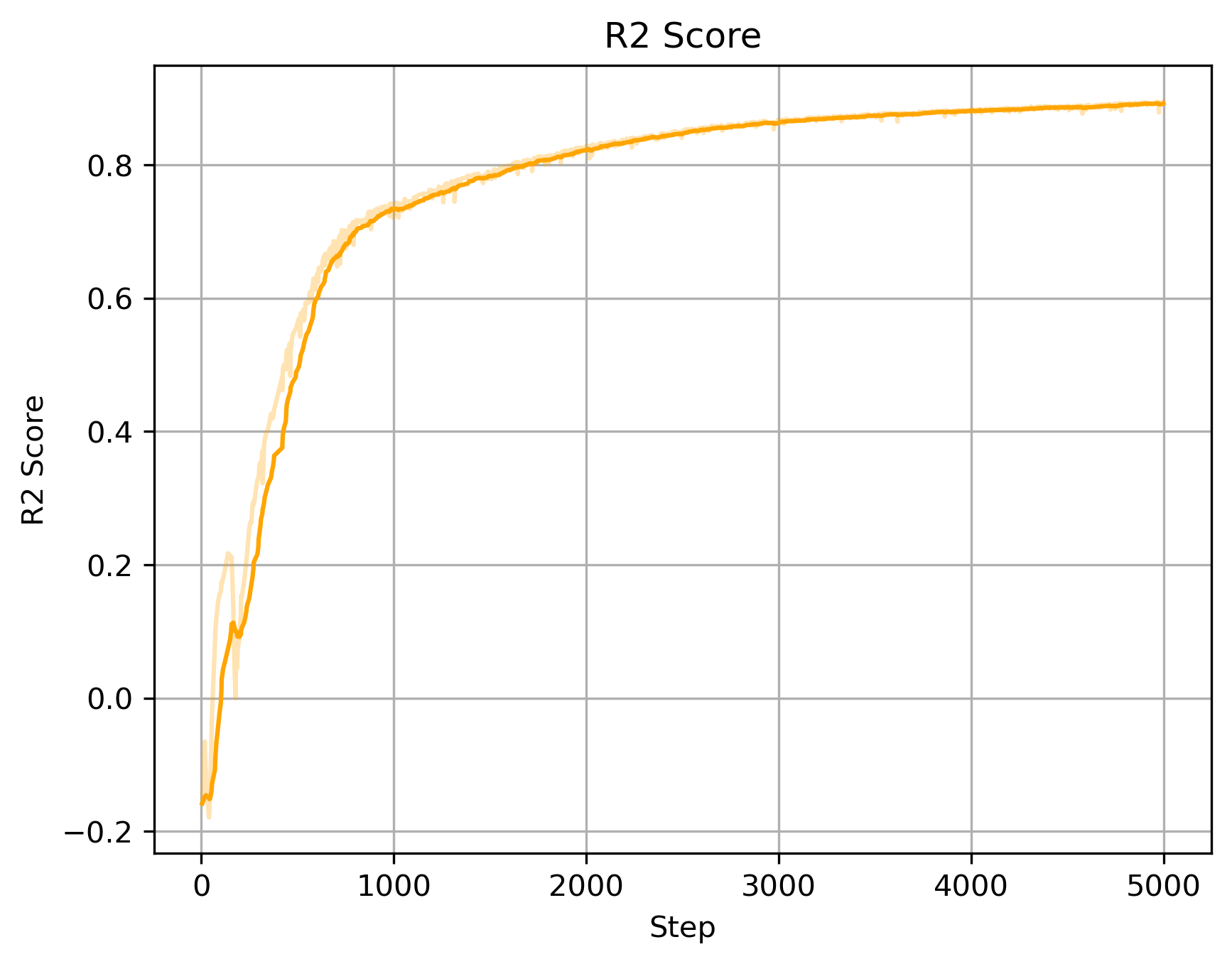}
    \caption{R2 score during training}
    \label{fig:R2_Score}
\end{figure}

Figures \ref{loss_plots} and \ref{fig:R2_Score} show that despite its architectural simplicity, this CNN implementation demonstrates promising results in predicting user interactions from particle positions alone.The training performance analysis of our CNN model for methane-nanotube threading reveals a comprehensive learning progression over 5000 training steps. The model demonstrates distinct learning behaviour throughout the training process, starting with negative R² values around -0.2, indicating initial predictions below baseline performance. The learning trajectory shows remarkable improvement in the early stages, with a particularly steep ascent during the first 2000 steps. During this period, the R² values rapidly improve from negative territory to approximately 0.8, demonstrating the model's growing capability to capture the underlying physics of methane-nanotube interactions.
As training progresses beyond the 2000-step mark, the learning curve exhibits a more gradual improvement, characteristic of fine-tuning behaviour. The model ultimately achieves a final R² score of approximately 0.87, indicating that it successfully explains 87\% of the variance in the interaction position and scale predictions. This high performance level suggests excellent capture of the essential features governing methane-nanotube threading dynamics.
The training curve's stability is particularly noteworthy, displaying consistent behaviour with minimal variance throughout the learning process. The narrow confidence band surrounding the main training curve indicates robust feature extraction. This stability, combined with the high final performance metrics, suggests that the model has successfully learned to predict the complex interactions between methane molecules and carbon nanotubes, making it a valuable tool for MD applications.
Several avenues for architectural enhancement should be explored:
\begin{itemize}
\item Integration of physics-informed constraints in the network design
\item Incorporation of temporal dependencies between consecutive frames
\item Extension to graph-based architectures to better capture molecular topology
\item Implementation of attention mechanisms for focusing on relevant atomic interactions
\end{itemize}
These enhancements could lead to more robust and generalizable models for predicting molecular interactions in virtual reality simulations, potentially improving the accuracy and efficiency of MD studies.

\section{Discussion and future directions}

 In this paper our results show that an agent's "intelligence" can be derived from an IL model trained on datasets generated by human experts within iMD-VR environment. When embodied as an avatar and reintegrated into the iMD-VR space, the agent's primary function is currently to clone the experts' behaviour by mimicking their actions which it successfully does. Judging by the current state of literature and the advancements made to IL and iMD-VR, and by combining these approaches with MAS IL approaches the agent's function may evolve beyond simple task replication. The focus shifts to the dynamics of human-agent interaction, creating a collaborative partner for research or training. For a teaching system, this avatar would not only demonstrate expertise by generating viable pathways for molecular interactions but would also need to engage with the human user in a natural and responsive manner. The success of this integration hinges on the agent's ability to model the complex, nuanced behaviours of a dyadic interaction, making the virtual collaboration feel seamless and intuitive, thereby effectively augmenting the human user's capabilities.
In this section, we identify two possible domains for the application of IL for iMD-VR: ligand/drug binding to proteins and material properties investigation. We also discuss the prospect of employing citizen science approaches to create datasets for training IL models.

\subsection{Ligand/drug binding to proteins}

One potential application of IL for iMD-VR is in computer-aided drug design (CADD) \cite{Walters2022}. CADD methods are being used within drug development to increase the efficiency of discovery, development and analysis of drug candidates \cite{sabe_current_2021}. Where the target protein structure is known, MD simulations can be used to shortlist candidate molecules for potential bioactivity by calculating protein-ligand complex stability \cite{shaker_silico_2021}. However, simulating ligand binding using MD remains a challenge due to the long simulation timescales required to simulate these rare events, and the associated computational cost.

iMD-VR provides a tool to address the challenge of simulating ligand binding: by harnessing the innate human ability to perform spatial tasks, iMD-VR enables researchers to induce complex binding events on accessible timescales. For example, Deeks et al.\cite{Deeks2020Interactive} demonstrated the use of iMD-VR for docking ligands to the main protease of the SARS-CoV-2 virus (the virus responsible for the COVID-19 pandemic). The authors found that iMD-VR experts were able to form docked structures that were in agreement with the crystal structures found experimentally. Another notable study by Deeks et al.\cite{Deeks2020} found that non-experts could also generate accurate structures of protein-ligand complexes. The results of these studies suggest that IL models could be trained effectively using data gathered from both expert and non-expert users, increasing the size of training sets that leverage human intuition to further sample these non-trivial rare events. Additionally, if trained on physically relevant trajectories, IL could be used to efficiently sample the space of relevant binding pathways, further enhancing the use of iMD-VR in the context of protein-ligand binding.

\subsection{Material properties investigation}

Another potential application of IL with iMD-VR is in the field of material design. Crossley-Lewis et al. \cite{Crossley-Lewis2023-jk} demonstrated the utility of iMD-VR for materials science, with emphasis on its research applications in fast-ion conduction and catalysis. The authors examined the defect and transport properties of the fast-ion conductor \ce{Li2O}---a promising energy storage material \cite{kondori2023lithiumoxide}---showing that the user interaction facilitated by iMD-VR enabled the researchers to investigate the mechanisms of ion transport rapidly without introducing significant bias towards unphysical regions of the potential energy landscape. If trained on such physically relevant data, an IL model could learn to optimize appropriate quantitative metrics that determine the likelihood of a given path, enabling efficient honing of pertinent mechanistic pathways to better understand the behaviour of fast-ion conductors, thus accelerating rational solid electrolyte design. 

Crossley-Lewis et al. \cite{Crossley-Lewis2023-jk} also used iMD-VR to examine the transport of the catalytic promoter methyl $n$-hexanoate through the H-ZSM-5 zeolite. Here, the researchers used iMD-VR to sample the dynamics of the promoter-zeolite system after the rare event of desorption of methyl $n$-hexanoate from a Br{\o}nsted acid site within the zeolite framework. By applying biasing forces to desorb the promoter molecule and guide it into various positions in the zeolite framework, the researchers were able to (a) investigate the transport dynamics of methyl $n$-hexanoate on accessible timescales, and (b) identify features of the zeolite structure relevant to the dynamics of the molecule. This study could be built upon by performing further quantitative analysis of both the energetics of desorption and the rate of diffusion of the promoter after desorption. An interesting research question is whether IL could enhance the sampling of relevant desorption pathways and desorbed promoter conformations. If successful, this protocol could be integrated into a typical MD workflow to achieve more efficient approximation of quantities of interest such as the diffusion coefficient---a measure of the average promoter diffusion, which greatly influences the catalytic efficiency of the material in question---helping to guide the search for effective catalysts and promoters.

\subsection{Citizen science for dataset generation}

In order for an ML model to perform well in a defined domain, it must be trained using high quality data. iMD-VR has previously been used to collect data via `citizen science' approaches, in which data collection is outsourced to non-expert users.
Deeks et al. \cite{Deeks2020} reported that novice iMD-VR users, many of whom were also not experts in ligand binding, could reliably reproduce experimentally-derived docking poses of flexible ligands with only a short amount of training ($<40$ minutes in VR). In another study, Shannon et al. \cite{Shannon2021} demonstrated that non-experts were able to explore the chemical reactivity of propyne with an OH radical, successfully identifying almost all of the important reaction pathways available to the system. Both of these studies indicate that in certain contexts non-chemist citizen scientists can generate chemically relevant datasets. This suggests that data collection need not be limited to experts in computational chemistry and/or iMD-VR. Furthermore, gamification techniques could be employed for data collection to increase participant engagement and reduce the need for bespoke in-person training. Such techniques have already been successfully applied in iMD-VR by Shannon et al. \cite{Shannon2021} (see above) and Roebuck Williams et al. \cite{roebuckWilliams2024}, who showed that non-expert citizen scientists could sense the properties of interactive molecular simulations in VR. 
 

\section{Conclusion}

This work has explored the potential of training AI agents using imitation learning on the rich dataset generated from iMD-VR. The immersive nature of VR environments enables researchers to apply their spatial reasoning abilities and domain expertise directly to molecular manipulation tasks, creating valuable demonstration data that captures human intuition and problem-solving strategies. Our proof-of-principle implementing a CNN BC model shows that the data structures in NanoVer can be used to train IL models capable of learning molecular interactions and solving molecular tasks. Drawing parallels from robotics, where IL has proven successful in teaching complex manipulation tasks, we believe that similar techniques can be adapted for molecular systems. Specifically, BC and IRL approaches could be particularly relevant for learning policies that guide molecular interactions in iMD-VR. The unique three-dimensional interaction data captured from expert demonstrations in iMD-VR systems offers unprecedented opportunities to encode domain knowledge into AI decision-making processes. However, several technical challenges remain, including the ensurance of generalization across diverse molecular systems, and handling the high dimensionality of conformational spaces. Future research directions should explore hybrid approaches that combine IL with reinforcement learning techniques, potentially leveraging recent advances in multi-agent systems and adversarial training methods. The integration of these approaches could lead to more robust and adaptable AI agents capable of assisting researchers in exploring vast conformational spaces, ultimately accelerating drug discovery, protein engineering, and materials design processes.
\backmatter
\section*{Declarations}

\bmhead{Competing interests}
On behalf of all authors, the corresponding author states that there is no conflict of interest.

\bmhead{Funding Information}
This work has been supported by the European Research Council under the European Union’s Horizon 2020 research and innovation programme through consolidator Grant NANOVR 866559, the Xunta de Galicia (Centro de investigación de Galicia accreditation), and the European Union (European Regional Development Fund—ERDF). DRG also thanks the Axencia Galega de Innovación for funding as an Investigador Distinguido through the Oportunius Program.

\bmhead{Author contribution}
All authors contributed to the study conception and design. Material and software preparation was done by J.B. Data collection, model development, and results analysis were performed by M.D. The first draft of the manuscript was written by M.D., R.R.W, and H.J.S. All authors reviewed, commented, and provided feedback on all versions of the manuscript. D.R.G. conceptualized the project, contributed to the final version of the manuscript, and supervised the project. All authors read and approved the final manuscript.

\bmhead{Data Availability Statement}
The data that support the findings of this study will be available from the corresponding author upon reasonable request.

\bmhead{Research Involving Human and/or Animals}
Not Applicable.

\bmhead{Informed Consent}
Not Applicable


\bibliography{ref_opt}

\end{document}